\documentclass[letterpaper, 10 pt, conference]{ieeeconf} 

\IEEEoverridecommandlockouts 

\usepackage{cite}
\usepackage{times} 
\usepackage{ctable}
\usepackage{epsfig} 
\usepackage{amsmath} 
\usepackage{amssymb} 
\usepackage{ragged2e}
\usepackage{amsfonts} 
\usepackage{pdfpages} 
\usepackage{multirow} 
\usepackage{algorithm2e}
\usepackage{graphicx,xcolor,colortbl} 
\usepackage[font=footnotesize]{caption} 

\def\ours{WLST}
\def\ourfusionstrategy{\textit{consistency fusion strategy}}
\def\detector{$M_{det}$}
\def\autolabeler{$M_{aut}$}
\definecolor{LightCyan}{rgb}{0.88,1,1}

\title{\LARGE \bf \ours{}: Weak Labels Guided Self-training for Weakly-supervised Domain Adaptation on 3D Object Detection}

\author{Tsung-Lin Tsou$^{1}$, Tsung-Han Wu$^{1}$, and Winston H. Hsu$^{1, 2}$
\thanks{$^{1}$National Taiwan University, $^{2}$Mobile Drive Technology}
}

\begin{document}

\maketitle
\thispagestyle{empty}
\pagestyle{empty}


\begin{abstract}

In the field of domain adaptation (DA) on 3D object detection, most of the work is dedicated to unsupervised domain adaptation (UDA). Yet, without any target annotations, the performance gap between the UDA approaches and the fully-supervised approach is still noticeable, which is impractical for real-world applications. On the other hand, weakly-supervised domain adaptation (WDA) is an underexplored yet practical task that only requires few labeling effort on the target domain. To improve the DA performance in a cost-effective way, we propose a general weak labels guided self-training framework, \ours{}, designed for WDA on 3D object detection. By incorporating autolabeler, which can generate 3D pseudo labels from 2D bounding boxes, into the existing self-training pipeline, our method is able to generate more robust and consistent pseudo labels that would benefit the training process on the target domain. Extensive experiments demonstrate the effectiveness, robustness, and detector-agnosticism of our \ours{} framework. Notably, it outperforms previous state-of-the-art methods on all evaluation tasks. Code and models are available at \url{https://github.com/jacky121298/WLST}.

\end{abstract}
\section{Introduction} \label{sec:Introduction}

\begin{figure*}[t]
    \centering
    \includegraphics[width=\linewidth]{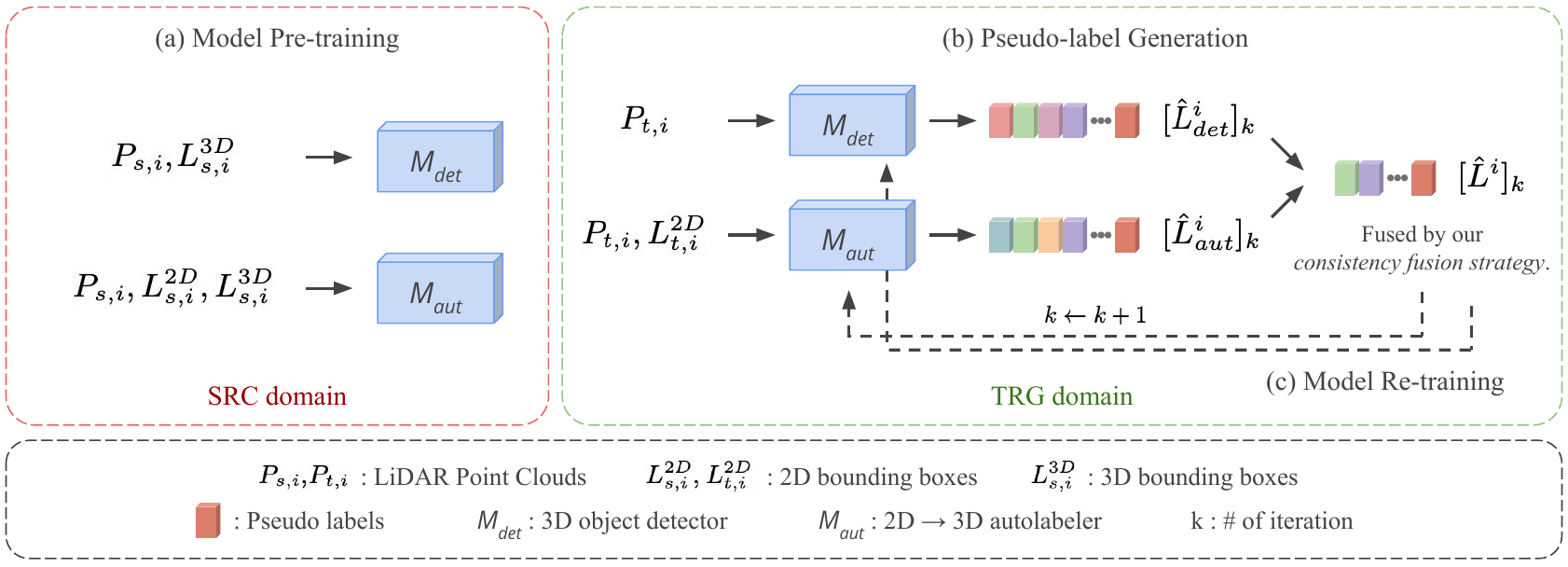}
    \caption{\textbf{Our \ours{} framework} is composed of three stages. (a) Pre-train 3D detector and autolabeler on the source data. (see Sec.~\ref{subsubsec:Model Pre-training}) (b) Generate high-quality pseudo labels by our \ourfusionstrategy{} on the target data. (see Sec.~\ref{subsubsec:Pseudo-label Generation}) (c) Re-train 3D detector and autolabeler on the pseudo-labeled target data. (see Sec.~\ref{subsubsec:Model Re-training})}
    \label{fig:framework}
\end{figure*}
\vspace{-4pt}

With the rapid development of 3D range sensors (\textit{e.g.} LiDAR point clouds) and large-scale human-annotated datasets \cite{caesar2020nuscenes, geiger2012we, sun2020scalability}, 3D object detection in the field of autonomous driving has garnered great attention and obtained remarkable breakthroughs \cite{lang2019pointpillars, shi2020pv, shi2022pv, shi2019pointrcnn, yan2018second, yin2021center, zhou2018voxelnet}. In order to deploy to real roads, 3D detectors must adapt to various real-world scenarios and perform robustly against numerous domain shifts arising from different settings of 3D range sensors, fickle weather conditions, miscellaneous objects in the driving scene, etc. However, existing 3D detectors are inadequate to tackle the domain gap realistically. Past work \cite{wang2020train} has shown that the performance of a fully-supervised 3D detector trained on Waymo Open Dataset \cite{sun2020scalability} dropped drastically when evaluated on KITTI Benchmark Dataset \cite{geiger2012we}. Therefore, developing an effective \textit{domain adaptation (DA)} approach is needed.

In the field of DA on 3D object detection, most of the work \cite{luo2021unsupervised, yang2021st3d, yang2022st3d++, you2022exploiting} is dedicated to \textit{unsupervised domain adaptation (UDA)}. Among them, the self-training approaches \cite{yang2021st3d, yang2022st3d++} perform the best. They redesigned the naive self-training pipeline to improve the pseudo-label selection mechanism and utilize effective augmentation techniques in the model training process, which achieved state-of-the-art performance in many DA tasks. Yet, without any target annotations, the performance gap between the UDA approaches (\textit{e.g.} 64.75 $\mbox{AP}_{\mbox{\tiny 3D}}$ on the Waymo $\rightarrow$ KITTI task) and the fully-supervised oracle approach (\textit{e.g.} 83.00 $\mbox{AP}_{\mbox{\tiny 3D}}$ in the KITTI dataset) is still noticeable as shown in Tab.~\ref{tab:Experimental Results}. On the other hand, few work has been contributed to \textit{weakly-supervised domain adaptation (WDA)}, among which SN \cite{wang2020train} utilizes object size statistics of the target domain to mitigate the domain shifts. However, its effectiveness largely depends on object size distributions and performs even worse than the UDA approaches (\textit{e.g.} 62.54 $\mbox{AP}_{\mbox{\tiny 3D}}$ on the Waymo $\rightarrow$ KITTI task). In summary, the above approaches are impractical for real-world applications.

To reduce such performance gaps in a cost-effective way, we propose a general weak labels guided self-training framework, \ours{}, designed for WDA on 3D object detection. Building upon the success of self-training UDA approaches \cite{yang2021st3d, yang2022st3d++} and studies \cite{liu2022map, wei2021fgr} on autolabeler that can generate 3D pseudo labels from 2D bounding boxes, our \ours{} framework incorporates autolabeler into the existing self-training pipeline. Specifically, as shown in Fig.~\ref{fig:framework}, a 3D detector and an autolabeler are first pre-trained on the labeled source domain. Then, pseudo labels would be generated by both models on the weak-labeled target domain. Finally, the 3D detector and autolabeler are iteratively improved by alternatively conducting pseudo-label generation and model re-training on these pseudo-labeled target data. Regarding annotation cost, statistics show that the time spent on annotating weak labels (\textit{i.e.} 2D bounding boxes) can be approximately three to sixteen times less than strong labels (\textit{i.e.} 3D bounding boxes) depending on the annotation tool used \cite{tang2019transferable}, rendering the cost affordable.

\begin{figure}[t]
    \centering
    \includegraphics[width=\linewidth]{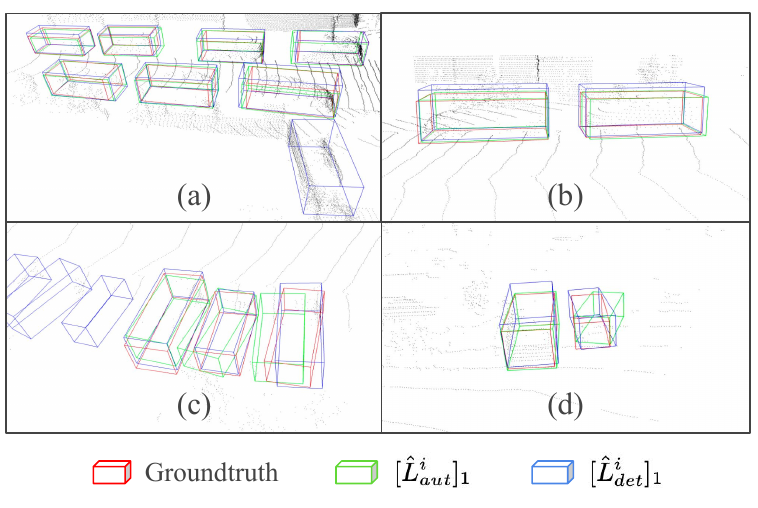}
    \caption{Visualization of pseudo labels $[\hat{L}_{det}^i]_1$ and $[\hat{L}_{aut}^i]_1$ generated by 3D detector and autolabeler respectively. We observed that \textbf{Top:} $[\hat{L}_{aut}^i]_1$ has higher precision. (a) It is less likely to predict extra FP boxes. (b) It is able to predict the heights of objects more precisely. \textbf{Bottom:} $[\hat{L}_{det}^i]_1$ has higher recall. (c, d) It has a better understanding of the correlation between objects, \textit{e.g.} a line of vehicles.}
    \label{fig:analysis}
\end{figure}

To further enhance the quality of pseudo labels generated by 3D detector and autolabeler, we design a pseudo-label selection mechanism to explore and leverage their distinct pros and cons. To elaborate, autolabeler has higher precision attributed to the fact that 2D bounding boxes help constrain the 3D search space for the pseudo labels as described in Fig.~\ref{fig:motivation}. Nevertheless, it works on the object level and couldn't learn the correlation between objects. On the other hand, 3D detector works on the scene level and has a larger Field of View (FoV), which enables a better understanding of the correlation between objects that leads to higher recall (see Fig.~\ref{fig:analysis}). Based on these observations, our proposed \ourfusionstrategy{} leverages geometric consistency and cross-modality consistency of pseudo labels to retain high precision and high recall simultaneously (see Tab.~\ref{tab:Pseudo-label Analysis}).

Extensive experiments on three widely used 3D object detection datasets, nuSenses Dataset \cite{caesar2020nuscenes}, KITTI Benchmark Dataset \cite{geiger2012we}, and Waymo Open Dataset \cite{sun2020scalability} demonstrate the effectiveness, robustness, and detector-agnosticism of our \ours{} framework. It can effectively close the performance gap between source only approach and fully-supervised oracle approach by up to 87.26$\%$ in $\mbox{AP}_{\mbox{\scriptsize BEV}}$ and up to 91.34$\%$ in $\mbox{AP}_{\mbox{\scriptsize 3D}}$ in Tab.~\ref{tab:Experimental Results}. Notably, we outperform previous state-of-the-art methods on all evaluation tasks. In summary, our main contributions are threefold:
\begin{itemize}
    \item We formulate and investigate the problem of WDA on 3D object detection, an underexplored yet practical task that has the potential to improve the DA performance in a cost-effective way.
    \item We propose \ours{}, a general weak labels guided self-training framework, to obtain more robust and consistent pseudo labels. To the best of our knowledge, we are the first to incorporate autolabeler into the self-training pipeline.
    \item Our \ours{} framework is extensively evaluated on three widely used 3D object detection datasets and outperforms previous state-of-the-art methods on all evaluation tasks.
\end{itemize}
\section{Related Work} \label{sec:Related Work}

\textbf{LiDAR-based 3D Object Detection.} Given the point clouds obtained from LiDAR sensors, 3D detectors aim to recognize and determine the 3D information of the objects, including location, dimension, orientation, and category. Based on data representations, 3D detectors can be divided into point-based, grid-based, and point-voxel-based. Point-based detectors \cite{qi2017pointnet, qi2017pointnet++, shi2019pointrcnn, yang2019std} first sample the point clouds and learn the features from gradually downsampled features. Grid-based detectors \cite{deng2021voxel, ge2020afdet, shi2020points, yin2021center, zheng2021cia, zhou2018voxelnet} first voxelize the point clouds into equally spaced grids and learn the features from these discrete grids. Point-voxel-based detectors \cite{shi2020pv, shi2022pv} utilize both points and voxels for 3D detection. In this work, we adopt PV-RCNN \cite{shi2020pv} as our 3D object detector.

\textbf{Domain Adaptation for 3D Detection.} Domain adaptation approaches aim to adapt the model trained on the source domain to the target domain. Wang \textit{et al.}~\cite{wang2020train} identify the difference in object size statistics as the key factor of domain shifts and normalize the object size distribution of the source domain by using its statistics of the target domain to mitigate the domain shifts. However, its effectiveness largely depends on object size distributions. MLC-Net \cite{luo2021unsupervised} leverages the teacher-student paradigm for pseudo-label generation via three levels of consistency to implement domain adaptation. Yang \textit{et al.}~\cite{yang2021st3d} further conclude that the domain shifts arise not only from the object size statistics but also from the point cloud distribution. They propose a new self-training pipeline called ST3D and achieve state-of-the-art performance in many DA tasks. Yet, without any target annotations, the performance gap between the UDA approach and the fully-supervised oracle approach is still noticeable. Therefore, we propose \ours{}, a general weak labels guided self-training framework to obtain more consistent pseudo labels and improve the DA performance as illustrated in Sec.~\ref{subsec:Weak Labels Guided Self-training Framework}.

\textbf{Weakly-supervised 3D Detection.} Weakly-supervised learning is a promising approach to utilize noisy, limited, or imprecise data to provide supervision signals and lessen the annotation cost. For 3D detection, weakly-supervised 3D approaches aim to obtain an autolabeler to enhance the weak labels into stronger forms (\textit{e.g.} from 2D bounding boxes to 3D boxes). Then, a 3D detector would be trained on these 3D pseudo labels. For example, Wei \textit{et al.}~\cite{wei2021fgr} propose a non-training frustum-aware geometric reasoning framework (FGR) to generate 3D pseudo labels from the frustum point clouds based on a 2D bounding box. Meng \textit{et al.}~\cite{meng2021towards} develop a quick BEV center click annotation strategy and generate 3D pseudo labels from these BEV center click annotations. Liu \textit{et al.}~\cite{liu2022map} introduce a trainable model called MAP-Gen, which leverages dense image information to tackle the sparsity issue of 3D point clouds and generates high-quality 3D pseudo labels from 2D bounding boxes. In spite of the state-of-the-art performance of MAP-Gen \cite{liu2022map}, it still needs a small amount of ground truth 3D labels to train its autolabeler. Despite the promising results obtained from the above methods, they fail to consider cross-domain scenarios. Hence, we propose an autolabeler designed for DA as illustrated in Sec.~\ref{subsubsec:Autolabeler}.
\section{Methodology} \label{sec:Methodology}

\begin{figure}[t]
    \centering
    \includegraphics[width=\linewidth]{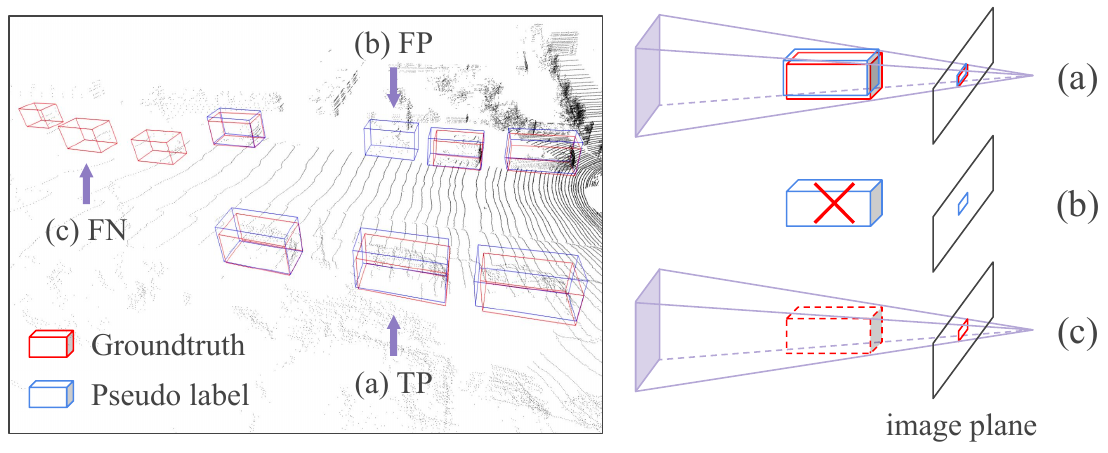}
    \caption{\textbf{Left:} Visualization of false positive (FP), false negative (FN), and true positive (TP) boxes of the pseudo labels. \textbf{Right:} According to the projective geometry, frustums can be generated by utilizing their 2D bounding boxes as the projection source and they define the 3D search space for pseudo labels, which manifests that an object should be located in the frustum corresponding to its 2D bounding box. In other words, when we re-project the pseudo labels into 2D image plane, (a) A TP box tends to have a higher IoU with its corresponding 2D bounding box. (b) A FP box does not have corresponding 2D bounding box and it is less likely to have a decent IoU with any 2D bounding box. (c) We can also learn that an object should exist in the frustum corresponding to a FN box.}
    \label{fig:motivation}
\end{figure}

We formulate the problem of weakly-supervised domain adaptation (WDA) on 3D object detection in Sec.~\ref{subsec:Problem Formulation} and present our weak labels guided self-training framework, \ours{}, in Sec.~\ref{subsec:Weak Labels Guided Self-training Framework}.


\subsection{Problem Formulation} \label{subsec:Problem Formulation}

Under the weakly-supervised domain adaptation (WDA) on 3D object detection setting, the goal is to adapt a 3D object detector from the labeled source domain $D_s = \{(P_{s,i}, L_{s,i}^{2D}, L_{s,i}^{3D})\}_{i=1}^{n_s}$ to the weak-labeled target domain $D_t = \{(P_{t,i}, L_{t,i}^{2D})\}_{i=1}^{n_t}$, where $n_s$ and $n_t$ denote the number of samples from the source and target domain respectively. Here, $P_{s,i}$, $L_{s,i}^{2D}$, and $L_{s,i}^{3D}$ represent LiDAR point clouds, 2D bounding boxes (\textit{i.e.} weak labels), and 3D bounding boxes (\textit{i.e.} strong labels) from the $i$-th source domain sample. The 2D bounding boxes are parameterized by their coordinates of the top-left and bottom-right corners in the image plane. (Different datasets might parameterize 2D bounding boxes differently.) The 3D bounding boxes are parameterized by their center location $(c_x,c_y,c_z)$, dimension $(l,w,h)$, orientation $\theta$, and category. Similarly, $P_{t,i}$ and $L_{t,i}^{2D}$ represent LiDAR point clouds and 2D bounding boxes from the $i$-th target domain sample.


\subsection{Weak Labels Guided Self-training Framework} \label{subsec:Weak Labels Guided Self-training Framework}

In this section, we present \ours{}, a general weak labels guided self-training framework that adapts the 3D detector trained on the labeled source domain to the weak-labeled target domain with the guidance of weak labels, which is shown in Fig.~\ref{fig:framework}. Our framework is composed of three stages: (1) Pre-train a 3D detector and an autolabeler on the source data (see Sec.~\ref{subsubsec:Model Pre-training}). (2) Generate high-quality pseudo labels by our \ourfusionstrategy{} on the target data (see Sec.~\ref{subsubsec:Pseudo-label Generation}). (3) Re-train the 3D detector and autolabeler on the pseudo-labeled target data (see Sec.~\ref{subsubsec:Model Re-training}).

\subsubsection{Autolabeler} \label{subsubsec:Autolabeler}

\begin{figure}[t]
    \centering
    \includegraphics[width=\linewidth]{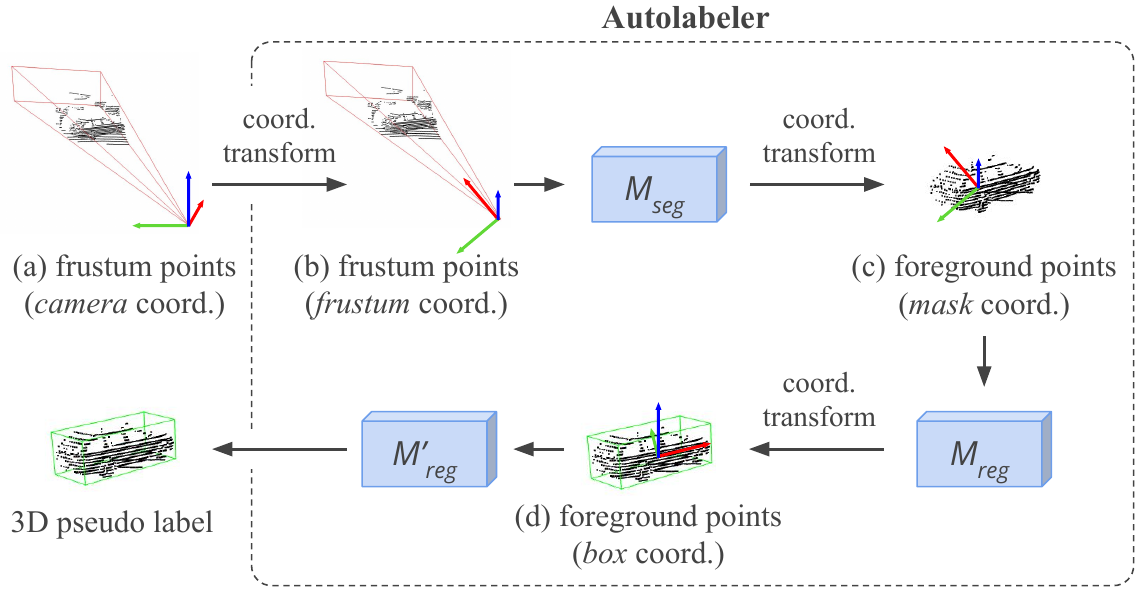}
    \caption{\textbf{Our proposed autolabeler designed for DA.} The model takes the frustum points in the camera coordinate as input and outputs a 3D pseudo label. ($M_{seg}$ denotes foreground segmentation network, and $M_{reg}$ denotes box regression network.)}
    \label{fig:autolabeler}
\end{figure}

An autolabeler aims to generate 3D pseudo labels from weak labels (\textit{i.e.} 2D bounding boxes). Despite the promising results obtained from \cite{liu2022map, wei2021fgr}, they fail to consider cross-domain scenarios. Hence, we propose an autolabeler designed for DA as illustrated in Fig.~\ref{fig:autolabeler}. Inspired by Frustum PointNets \cite{qi2018frustum} and Cascade-RCNN \cite{cai2018cascade}, we adopt coordinate transformations (\textit{e.g.} frustum coordinate, mask coordinate) to canonicalize the point cloud for more effective learning and utilize cascaded box regression networks to fine-tune the pseudo boxes iteratively.

Specifically, we first extract the frustum points from a given 2D bounding box in the camera coordinate shown in Fig.~\ref{fig:autolabeler} (a). With the known camera intrinsic and extrinsic matrices, the frustum can be generated by utilizing a 2D bounding box as the projection source and it defines a 3D search space for the pseudo label. We then gather the point clouds within the frustum to form frustum points as the input of autolabeler.

To make the distribution of frustum points more aligned across objects, we transform their coordination to orthogonalize the +X axis of the frustum to the image plane shown in Fig.~\ref{fig:autolabeler} (b). Then, the frustum points are passed to a PointNet \cite{qi2017pointnet} based foreground segmentation network $M_{seg}$ to extract foreground points. Furthermore, in order to perform residual-based 3D localization, foreground points are transformed to the mask coordinate by translating their centroid to the origin shown in Fig.~\ref{fig:autolabeler} (c).

Subsequently, to perform robustly against numerous domain shifts, we utilize two cascaded PointNet \cite{qi2017pointnet} based box regression networks $M_{reg}$ and $M'_{reg}$ to regress the 3D pseudo label. To be specific, we utilize the first network $M_{reg}$ to predict the initial 3D bounding box in the first place. Then, foreground points are transformed to the box coordinate, in which the box’s orientation is parallel to the +X axis and the box’s center is at the origin shown in Fig.~\ref{fig:autolabeler} (d), to perform residual-based 3D localization and orientation. Ultimately, we generate the final 3D pseudo label from the second network $M'_{reg}$.


\subsubsection{Model Pre-training} \label{subsubsec:Model Pre-training}

Our \ours{} framework starts from pre-training a 3D detector and an autolabeler on the labeled source domain $D_s = \{(P_{s,i}, L_{s,i}^{2D}, L_{s,i}^{3D})\}_{i=1}^{n_s}$. Apart from valuable knowledge, the pre-trained models also learn the biases from the source domain due to inevitable domain shifts. According to recent works \cite{luo2021unsupervised, wang2020train, yang2021st3d} which aim to mitigate the domain shifts, they unanimously agreed that the bias in object size statistics has harmful impacts on 3D object detection and leads to inaccurate size prediction of 3D bounding boxes on the target domain. To alleviate this problem, we randomly rescale the object size similar to \cite{yang2021st3d} in the pre-training process to simulate the diverse object size distribution on the target domain, which makes the 3D detector and autolabeler more robust against object size bias.

\subsubsection{Pseudo-label Generation} \label{subsubsec:Pseudo-label Generation}

With the pre-trained 3D detector and autolabeler, the next step is to generate pseudo labels on the weak-labeled target domain. For clarity, we refer to $[\hat{L}_{det}^i]_k$ as initial pseudo labels generated by 3D detector \detector{} at the $k$-th iteration and $[\hat{L}_{aut}^i]_k$ as initial pseudo labels generated by autolabeler \autolabeler{} at the $k$-th iteration. Note that non-maximum suppression (NMS) was conducted for $[\hat{L}_{det}^i]_k$ to get rid of the redundant boxes.

\textbf{Consistency Fusion Strategy.} We propose \ourfusionstrategy{} to effectively select high-quality pseudo labels $[\hat{L}^i]_k$ from $[\hat{L}_{det}^i]_k$ and $[\hat{L}_{aut}^i]_k$ in accordance with \textit{geometric consistency} and \textit{cross-modality consistency} of these pseudo labels.

For \textit{geometric consistency}, we propose a 2D Intersection over Union (IoU) based criterion to assess the existence probability of pseudo labels. Specifically, as illustrated in Fig.~\ref{fig:motivation}, objects should be located in the frustums corresponding to their 2D bounding boxes. In other words, when we re-project the pseudo labels into 2D image plane with the known camera projection matrix, a TP box tends to have a higher IoU with its corresponding 2D bounding box, whereas a FP box is less likely to have a decent IoU with any 2D bounding box. To be more specific, for the pseudo label in $[\hat{L}_{aut}^i]_k$ which has an exact corresponding 2D bounding box, we calculate the 2D IoU between the convex hull of the re-projected pseudo label's corners and its corresponding 2D bounding box in the image plane as the existence probability of this pseudo label, which is denoted as $prob$. For the pseudo label in $[\hat{L}_{det}^i]_k$ which has no exact corresponding 2D bounding box, we first calculate the 2D IoU matrix $E = e_w \in \mathbb{R}^{n_w}$ between the convex hull of the re-projected pseudo label's corners and $n_w$ 2D bounding boxes in $L_{t,i}^{2D}$. Then, we take the maximum value in $E$ as the existence probability of this pseudo label. To the best of our knowledge, we are the first to demonstrate that it can be a good criterion to assess the existence probability of the pseudo labels.

For \textit{cross-modality consistency}, we match and fuse the pseudo labels generated by different modalities (\textit{i.e.} 3D detector and autolabeler) that have similar locations, dimensions, and orientations. To be more specific, we calculate the pair-wise 3D IoU matrix $I = {i_{j, j'}} \in \mathbb{R}^{n_u \times n_v}$ between each box in $[\hat{L}_{det}^i]_k$ and each box in $[\hat{L}_{aut}^i]_k$. Here, we assume that $[\hat{L}_{det}^i]_k$ contains $n_u$ boxes and is denoted as $[\hat{L}_{det}^i]_k = \{(box_u, s_u, prob_u)^k_j\}^{n_u}_{j=1}$, which are box parameters, predicted confidence score, and the existence probability of this box respectively. Similarly, we assume $[\hat{L}_{aut}^i]_k$ contains $n_v$ boxes and is denoted as $[\hat{L}_{aut}^i]_k = \{(box_v, s_v, prob_v)^k_{j'}\}^{n_v}_{j'=1}$. For all pair-wise boxes $(box_u, s_u, prob_u)^k_j$ and $(box_v, s_v, prob_v)^k_{j'}$, they are successfully matched if they achieve both \textit{geometric consistency} and \textit{cross-modality consistency} as
\begin{equation}
    \left\{
    \begin{array}{l}
        \max(prob_u, prob_v) \ge T_{exist}, \\
        i_{j, j'} > 0.1,
    \end{array} \right.
\end{equation}
Note that we set $T_{exist}=0.7$ refer to KITTI 2D object detection benchmark \cite{geiger2012we}. Later, they are further fused by only keeping the box with a higher confidence score and then cached into the $[\hat{L}^i]_k$ as $(box, s, prob)^k =$
\begin{equation}
    \left\{
    \begin{array}{lll}
        (box_u, s_u, prob_u)^k_j    & , & \mbox{if } s_u > s_v, \\[2pt]
        (box_v, s_v, prob_v)^k_{j'} & , & \mbox{otherwise},
    \end{array} \right.
\end{equation}
For other unmatched boxes in $[\hat{L}_{det}^i]_k$ and $[\hat{L}_{aut}^i]_k$, they fail to achieve either \textit{geometric consistency} or \textit{cross-modality consistency}. Hence, we lower their confidence scores by their existence probability due to their higher uncertainty as $s = s \times prob$ and then cached into the $[\hat{L}^i]_k$. Eventually, we filter out the ambiguous boxes whose confidence scores are lower than a threshold $T$. (We set $T=0.6$ in practice.) Benefited from our \ourfusionstrategy{}, we could generate more robust and consistent pseudo boxes to improve the process of model re-training.

\newcommand\ts{\rule{0pt}{1em}}
\newcommand\bs{\rule[-0.4em]{0pt}{0pt}}
\newcommand\tts{\rule{0pt}{0.0pt}}
\setlength{\tabcolsep}{0.04\linewidth}{
\begin{table*}[t]
    \centering
    \begin{tabular}{c|c|c|cc|cc}
        \specialrule{1pt}{0em}{0em}
        \small Task & \small Setting & \small Method & \small $\mbox{AP}_{\mbox{\scriptsize BEV}}$ & \small Closed Gap & \small $\mbox{AP}_{\mbox{\scriptsize 3D}}$ & \small Closed Gap \ts\bs\\\hline\hline
        \multirow{8}{*}{\small Waymo $\rightarrow$ KITTI} & - & Source Only & 60.32 & - & 21.66 & - \ts\bs\\\cline{2-7}
        & \multirow{2}{*}{UDA} & ST3D \cite{yang2021st3d} & 83.37 & +75.50\% & 64.75 & +70.25\% \ts\\
        & & $\mbox{ST3D++ \cite{yang2022st3d++}}^\dagger$ & 84.59 & +79.50\% & 67.73 & +75.11\% \bs\\\cline{2-7}
        & \multirow{4}{*}{WDA} & SN \cite{wang2020train} & 78.24 & +58.70\% & 62.54 & +66.64\% \ts\\
        & & ST3D (w/ SN) \cite{yang2021st3d} & 86.53 & +85.85\% & 76.85 & +89.97\% \tts\\
        & & $\mbox{ST3D++ (w/ SN) \cite{yang2022st3d++}}^\dagger$ & 86.92 & +87.13\% & 77.36 & +90.81\% \tts\\
        & & \cellcolor{LightCyan}\ours{} (Ours) & \cellcolor{LightCyan}\textbf{86.96} & \cellcolor{LightCyan}\textbf{+87.26\%} & \cellcolor{LightCyan}\textbf{77.69} & \cellcolor{LightCyan}\textbf{+91.34\%} \bs\\\cline{2-7}
        & - & Oracle &  90.85 & - & 83.00 & - \ts\bs\\
        \hline\hline
        \multirow{6}{*}{\small Waymo $\rightarrow$ nuScenes} & - & Source Only & 34.51 & - & 21.44 & - \ts\bs\\\cline{2-7}
        & UDA & ST3D \cite{yang2021st3d} & 36.38 & +9.99\% & 22.99 & +9.03\% \ts\bs\\\cline{2-7}
        & \multirow{3}{*}{WDA} & SN \cite{wang2020train} & 34.95 & +0.02\% & 22.19 & +4.37\% \ts\\
        & & ST3D (w/ SN) \cite{yang2021st3d} & 36.65 & +11.43\% & 23.66 & +12.93\% \tts\\
        & & \cellcolor{LightCyan}\ours{} (Ours) & \cellcolor{LightCyan}\textbf{39.54} & \cellcolor{LightCyan}\textbf{+26.87\%} & \cellcolor{LightCyan}\textbf{24.46} & \cellcolor{LightCyan}\textbf{+17.59\%} \bs\\\cline{2-7}
        & - & Oracle & 53.23 & - & 38.61 & - \ts\bs\\
        \hline\hline
        \multirow{6}{*}{\small nuScenes $\rightarrow$ KITTI} & - & Source Only & 69.26 & - & 39.17 & - \ts\bs\\\cline{2-7}
        & UDA & ST3D \cite{yang2021st3d} & 77.38 & +37.61\% & 70.86 & +72.30\% \ts\bs\\\cline{2-7}
        & \multirow{3}{*}{WDA} & SN \cite{wang2020train} & 60.12 & -42.33\% & 46.23 & +16.11\% \ts\\
        & & ST3D (w/ SN) \cite{yang2021st3d} & 83.84 & +67.53\% & 72.91 & +76.98\% \tts\\
        & & \cellcolor{LightCyan}\ours{} (Ours) & \cellcolor{LightCyan}\textbf{87.16} & \cellcolor{LightCyan}\textbf{+82.91\%} & \cellcolor{LightCyan}\textbf{77.73} & \cellcolor{LightCyan}\textbf{+87.98\%} \bs\\\cline{2-7}
        & - & Oracle & 90.85 & - & 83.00 & - \ts\bs\\
        \specialrule{1pt}{0em}{0em}
    \end{tabular}
    \caption{\textbf{Experiment results on three DA tasks.} Our \ours{} adopts PV-RCNN \cite{shi2020pv} as 3D detector and outperforms all existing methods on $\mbox{AP}_{\mbox{\scriptsize BEV}}$ and $\mbox{AP}_{\mbox{\scriptsize 3D}}$ of the car category at IoU = 0.7. The reported AP are the results on the moderate case when KITTI is regarded as the target domain and are the overall results for other DA tasks. We also report the closed gap to assess how much the performance gap between Source Only and Oracle is closed. $\dagger$~refers to the results reported by \cite{yang2022st3d++}.}
    \label{tab:Experimental Results}
\end{table*}
}


\subsubsection{Model Re-training} \label{subsubsec:Model Re-training}

With the high-quality pseudo labels $[\hat{L}^i]_k$ generated by our \ourfusionstrategy{}, we re-train the 3D detector and autolabeler on $\{(P_{t,i}, L_{t,i}^{2D}, [\hat{L}^i]_k\}$. Moreover, we use curriculum data augmentation (CDA) technique proposed by \cite{yang2021st3d} in the model re-training process to gradually generate more challenging cases for the benefit of training process.
\section{Experiments} \label{sec:Experiments}


\subsection{Experiment Settings} \label{subsec:Experiment Settings}

\textbf{Datasets.} Our experiments are conducted on three widely used 3D object detection datasets, nuSenses Dataset \cite{caesar2020nuscenes}, KITTI Benchmark Dataset \cite{geiger2012we}, and Waymo Open Dataset \cite{sun2020scalability}, and focus on three DA tasks: ($\romannumeral 1$) Waymo $\rightarrow$ KITTI, ($\romannumeral 2$) Waymo $\rightarrow$ nuScenes, and ($\romannumeral 3$) nuScenes $\rightarrow$ KITTI.


\textbf{Method Comparison.} We compare our \ours{} with other unsupervised approaches (\textit{i.e.} Source Only, ST3D \cite{yang2021st3d}, ST3D++ \cite{yang2022st3d++}), weakly-supervised approaches (\textit{i.e.} SN \cite{wang2020train}, ST3D (w/ SN) \cite{yang2021st3d}, ST3D++ (w/ SN) \cite{yang2022st3d++}), and fully-supervised approach (\textit{i.e.} Oracle). (1) \textit{Source Only} directly evaluates the source domain pre-trained model on the target domain. (2) \textit{SN} \cite{wang2020train} is a baseline WDA method that leverages object size statistics of the target domain. (3) \textit{ST3D} and \textit{ST3D++} are the state-of-the-art UDA approaches. (4) \textit{ST3D (w/ SN)} and \textit{ST3D++ (w/ SN)} are the state-of-the-art WDA approaches which are equipped with the SN. (5) \textit{Oracle} evaluates the fully-supervised model trained on the target domain.

\textbf{Evaluation Metric.} We follow \cite{yang2021st3d} and adopt the KITTI evaluation metric on the car category, which is known as the vehicle in the Waymo Open Dataset. In addition, we evaluate objects in the ring view except KITTI dataset as it only provides 2D and 3D bounding box annotations for objects within the Field of View (FoV) of the front camera. We report the Average Precision (AP) over 40 recall positions, and set the IoU thresholds as 0.7 for both the bird’s eye view (BEV) IoU and 3D IoU. We also use the closed gap evaluation metric proposed by \cite{yang2021st3d} to assess how much the performance gap between Source Only and Oracle is closed, which is $\textbf{closed gap} = {{\mbox{\small AP}_{\mbox{\scriptsize Method}} - \mbox{\small AP}_{\mbox{\scriptsize Source Only}}} \over {\mbox{\small AP}_{\mbox{\scriptsize Oracle}} - \mbox{\small AP}_{\mbox{\scriptsize Source Only}}}} \times \small 100\%$.



\setlength{\tabcolsep}{0.015\linewidth}{
\begin{table}[t]
    \centering
    \begin{tabular}{l|c|c}
        \specialrule{1pt}{0em}{0em}
        Method & Fusion & $\mbox{AP}_{\mbox{\scriptsize BEV}}$ / $\mbox{AP}_{\mbox{\scriptsize 3D}}$ \ts\bs\\\hline\hline
        3D detector only & & 80.97 / 64.53 \ts\\
        Autolabeler only & & 83.36 / 71.22 \bs\\\hline
        Non-Maximum Suppression (NMS) & \checkmark & 86.49 / 76.89 \ts\\
        Bayesian Fusion \cite{chen2022multimodal} & \checkmark & 86.29 / 76.39 \tts\\
        CLOCs3D \cite{pang2020clocs} & \checkmark & 85.75 / 75.92 \tts\\
        Consistency Fusion Strategy (Ours) & \checkmark & \textbf{89.14} / \textbf{77.69} \bs\\
        \specialrule{1pt}{0em}{0em}
    \end{tabular}
    \caption{\textbf{Fusion Strategy Analysis.} We compare our fusion strategy to other fusion strategies and report $\mbox{AP}_{\mbox{\scriptsize BEV}}$ and $\mbox{AP}_{\mbox{\scriptsize 3D}}$ of the car category at IoU = 0.7 on the Waymo $\rightarrow$ KITTI task. The results suggest that either directly using pseudo labels from 3D detector or autolabeler is suboptimal. In contrast, our proposed \ourfusionstrategy{} obtains the best outcome on both $\mbox{AP}_{\mbox{\scriptsize BEV}}$ and $\mbox{AP}_{\mbox{\scriptsize 3D}}$.}
    \label{tab:Fusion Strategy Analysis}
\end{table}
}


\subsection{Experiment Results} \label{subsec:Experiment Results}

We analyze the experiment results in terms of three DA scenarios: ($\romannumeral 1$) Waymo $\rightarrow$ KITTI: domains with a larger difference in object size statistics, ($\romannumeral 2$) Waymo $\rightarrow$ nuScenes: domains with a larger difference in point cloud distribution, and ($\romannumeral 3$) nuScenes $\rightarrow$ KITTI: domains with a larger difference in object size statistics as well as in point cloud distribution.

For the first scenario (\textit{i.e.} Waymo $\rightarrow$ KITTI), we found that it is a relatively simple task due to the fact that both domains have dense point cloud distribution by utilizing 64-beam LiDAR. Any method on this task can effectively close the performance gap between Source Only and Oracle. Yet, our method still outperforms all UDA methods by a large margin (around $\sim$2$\%$ in $\mbox{AP}_{\mbox{\scriptsize BEV}}$, $\sim$10$\%$ in $\mbox{AP}_{\mbox{\scriptsize 3D}}$) and better than the WDA methods by around $\sim$0.04$\%$ in $\mbox{AP}_{\mbox{\scriptsize BEV}}$ and around $\sim$0.3$\%$ in $\mbox{AP}_{\mbox{\scriptsize 3D}}$. These encouraging results validate that our method can effectively close the performance gap by 87.26$\%$ in $\mbox{AP}_{\mbox{\scriptsize BEV}}$ and 91.34$\%$ in $\mbox{AP}_{\mbox{\scriptsize 3D}}$.

For the second scenario (\textit{i.e.} Waymo $\rightarrow$ nuScenes), we found that it is a relatively difficult task when we adapt detectors from the domain with denser point cloud distribution (\textit{e.g.} 64-beam LiDAR) to the domain with sparser point cloud distribution (\textit{e.g.} 32-beam LiDAR). We observed that the baseline method SN only has minor performance gain when the domain shifts in object size statistics is subtle. However, our method also attains a considerable performance gain and outperforms all existing methods.

For the third scenario (\textit{i.e.} nuScenes $\rightarrow$ KITTI), despite its larger difference in point cloud distribution, we found it relatively easy to adapt detectors when the target domain has denser point cloud distribution. That is, it manifests that the point density of the target domain is more crucial on DA tasks than the point density of the source domain. We can obtain comparable performance on KITTI dataset regardless of the point density of the source domain (\textit{e.g.} Waymo $\rightarrow$ KITTI task, nuScenes $\rightarrow$ KITTI task). Furthermore, our method outperforms current state-of-the-art WDA method by a large margin (around $\sim$3$\%$ in $\mbox{AP}_{\mbox{\scriptsize BEV}}$, $\sim$5$\%$ in $\mbox{AP}_{\mbox{\scriptsize 3D}}$).

These promising results validate that our method can effectively adapt the 3D object detector trained on the source domain to the target domain and perform robustly against numerous domain shifts.

\newcommand\T{\rule{0pt}{2.5ex}}
\newcommand\B{\rule[-1ex]{0pt}{0pt}}

\setlength{\tabcolsep}{0.05\linewidth}{
\begin{table}[t]
    \centering
    \begin{tabular}{c|c|c}
        \specialrule{1pt}{0em}{0em}
        Pseudo Labels & Recall 0.7 & Precision 0.7 \\\hline\hline
        $[\hat{L}_{det}^i]_k$ & \textbf{50.38} & 69.11 \T\B\\\hline
        $[\hat{L}_{aut}^i]_k$ & 45.54 & 72.48 \T\B\\\hline
        $[\hat{L}^i]_k$ & 48.01 & \textbf{78.20} \T\B\\
        \specialrule{1pt}{0em}{0em}
    \end{tabular}
    \caption{\textbf{Qualitative analysis on pseudo labels.} We evaluate the quality of pseudo labels on the Waymo $\rightarrow$ KITTI task by Recall with IoU $>$ 0.7 and Precision with IoU $>$ 0.7. The pseudo labels $[\hat{L}_{det}^i]_k$, $[\hat{L}_{aut}^i]_k$, and $[\hat{L}^i]_k$ are generated by 3D detector, autolabeler, and later fused by our \ourfusionstrategy{} respectively. Our fusion strategy effectively eliminates the redundant FP boxes to obtain high precision and retain high recall simultaneously.}
    \label{tab:Pseudo-label Analysis}
\end{table}
}


\subsection{Ablation Studies} \label{subsec:Ablation Study}

\textbf{Fusion Strategy Analysis.} As demonstrated in Tab.~\ref{tab:Fusion Strategy Analysis}, we conduct fusion strategy analysis on the Waymo $\rightarrow$ KITTI task. Apart from our \ourfusionstrategy{}, we also study other fusion strategies like Non-Maximum Suppression (NMS), Bayesian Fusion \cite{chen2022multimodal}, and CLOCs3D. Bayesian Fusion \cite{chen2022multimodal} is a non-learning based fusion strategy derived from the Bayes’ rule that assumes conditional independence across modalities. CLOCs3D is extended from CLOCs \cite{pang2020clocs} and we modified the feature tensor in \cite{pang2020clocs} as $T_{i,j} = \{IoU_{i,j}^{3D}, s_i^{3D}, s_j^{3D}, prob_i, prob_j\}$ where $IoU_{i,j}^{3D}$ denotes 3D IoU between pseudo labels, $s$ denotes the predicted confidence score, and $prob$ denotes the existence probability of the pseudo label. Surprisingly, we found that the baseline strategy NMS performs well enough by only selecting boxes with higher confidence scores. Yet, the learning-based fusion strategy CLOCs3D does not perform well possibly because the large difference in the input data distribution (\textit{i.e.} feature tensor) from different domains affects its efficacy. In contrast, our \ourfusionstrategy{} performs the best as it leverages geometric consistency and cross-modality consistency to obtain more robust and consistent pseudo labels.

To further validate the effectiveness of our \ourfusionstrategy{}, we also conduct qualitative analysis on the pseudo labels $[\hat{L}_{det}^i]_k$, $[\hat{L}_{aut}^i]_k$, and $[\hat{L}^k]_k$ which are generated by 3D detector, autolabeler, and later fused by our \ourfusionstrategy{} respectively. According to the statistical results in Tab.~\ref{tab:Pseudo-label Analysis}, we found that $[\hat{L}_{aut}^i]_k$ has higher precision attributed to the fact that 2D bounding boxes help constrain the 3D search space for the pseudo labels as described in Fig.~\ref{fig:motivation}. $[\hat{L}_{det}^i]_k$ has higher recall as it has a larger Field of View (FoV), which enables a better understanding of the correlation between objects. Nevertheless, our \ourfusionstrategy{} effectively eliminates the redundant FP boxes to obtain high precision and retain high recall simultaneously.

\textbf{Component Analysis in Autolabeler.} We propose an autolabeler designed for DA as shown in Fig.~\ref{fig:autolabeler}. Inspired by Frustum PointNets \cite{qi2018frustum} and Cascade-RCNN \cite{cai2018cascade}, we adopt coordinate transformations (\textit{e.g.} frustum coordinate, mask coordinate) to canonicalize the point cloud for more effective learning and utilize cascaded box regression networks to fine-tune the pseudo boxes iteratively. As illustrated in Tab.~\ref{tab:Component Analysis in Autolabeler}, we see that both coordinate transformation components effectively make the distribution of points more aligned across objects and render the autolabeler converge easier. Moreover, the design of cascaded network further make the autolabeler perform robustly against domain shifts.



\setlength{\tabcolsep}{0.012\linewidth}{
\begin{table}[t]
    \centering
    \begin{tabular}{ccc|cc}
        \specialrule{1pt}{0em}{0em}
        \multicolumn{3}{c}{Components}\vline & \multirow{3}{*}{\footnotesize Recall 0.7} & \multirow{3}{*}{\footnotesize Precision 0.7} \\\cline{1-3}
        \shortstack{frustum coord. \\ transform} & \shortstack{mask coord. \\ transform} & \shortstack{cascaded \\ networks} & & \rule{0pt}{2em}\bs\\\hline\hline
        & & & 10.26 & 46.87 \ts\\
        \checkmark & & & 33.60 & 58.02 \tts\\
        \checkmark & \checkmark & & 34.07 & 60.84 \tts\\
        \checkmark & \checkmark & \checkmark & \textbf{45.54} & \textbf{72.48} \bs\\
        \specialrule{1pt}{0em}{0em}
    \end{tabular}
    \caption{\textbf{Component Analysis in Autolabeler.} We evaluate the quality of pseudo labels generated by autolabeler on the Waymo $\rightarrow$ KITTI task by Recall with IoU $>$ 0.7 and Precision with IoU $>$ 0.7. The results validate the effectiveness of the coordinate transformation components and the design of cascaded networks.}
    \label{tab:Component Analysis in Autolabeler}
\end{table}
}
\section{Conclusion} \label{sec:Conclusion}


We propose a general weak labels guided self-training framework, \ours{}, designed for WDA on 3D object detection. By incorporating autolabeler into the existing self-training pipeline, our method is able to generate more robust and consistent pseudo labels. Extensive experiments demonstrate the effectiveness of our framework.
\section{Acknowledgement}

This work was supported in part by National Science and Technology Council, Taiwan, under Grant NSTC 112-2634-F-002-006 and by Qualcomm through a Taiwan University Research Collaboration Project. We are grateful to Mobile Drive Technology Co., Ltd (MobileDrive) and the National Center for High-performance Computing.




\clearpage

\nocite{*}

\clearpage
\appendix


\subsection{Implementation Details} \label{subsec:Implementation Details}

In this section, we provide parameter setups for domain
adaptation tasks in Sec.~\ref{subsubsec:Parameter Setups} and present implementation details of autolabeler in Sec.~\ref{subsubsec:Implementation Details of Autolabeler}.

\subsubsection{Parameter Setups} \label{subsubsec:Parameter Setups}

We adopt PV-RCNN \cite{shi2020pv} as our 3D detector and propose a framework in Sec.~\ref{subsubsec:Autolabeler} as our autolabeler. For model pre-training on the source domain, we use Adam \cite{kingma2014adam} with learning rate $1 \times 10^{-3}$ to pre-train the autolabeler and adopt the training configurations of pointcloud-based object detection codebase OpenPCDet \cite{openpcdet2020} to pre-train the 3D detector. For model re-training on the target domain, we use Adam \cite{kingma2014adam} with learning rate $1 \times 10^{-3}$ and one cycle scheduler to fine-tune the autolabeler and 3D detector for 50 epochs. In addition, we update the pseudo labels with our \ourfusionstrategy{} for every 5 epochs.

We adopt the widely used data augmentation techniques during model pre-training and re-training processes such as random object scaling, random object rotation, random world scaling, random world rotation, and random world flipping. Moreover, we use curriculum data augmentation (CDA) technique proposed by \cite{yang2021st3d} in the model re-training process to gradually generate more challenging cases for the benefit of the training process. All datasets are transformed to the unified point cloud coordinate where the center of the coordinate system is shifted to the ground plane.

For all domain adaptation tasks, we pre-train the 3D detector for 50 epochs and the autolabeler for 100 epochs on the labeled source domain. Then, we re-train both models for 50 epochs on the weak-labeled target domain. The detection range is $[-75.2, 75.2] m$ for X and Y axes, and $[-2, 4] m$ for the Z axis. Moreover, we follow \cite{yang2021st3d} and set the voxel size of the 3D detector to $(0.1m, 0.1m, 0.15m)$ on all datasets for a fair evaluation.

\subsubsection{Implementation Details of Autolabeler} \label{subsubsec:Implementation Details of Autolabeler}

Our proposed autolabeler is composed of a foreground segmentation network and a box regression network as illustrated in the main paper.

The foreground segmentation network is a PointNet \cite{qi2017pointnet} based network, where each point cloud is first processed by five layers of multi-layer perceptron (MLP) and the output channel sizes are 64, 64, 64, 128, and 1024 respectively. In addition, every layer of the MLP contains batch normalization and ReLU. The last layer's output of per-point MLP (1024-dim) is pooled with a max pooling layer and concatenated with the second layer's output (64-dim) to form the 1088-dimension features. Then, the concatenated features are processed by five layers of MLP and the output channel sizes are 512, 256, 128, 128, and 2 respectively. Moreover, the last layer does not have batch normalization or ReLU. Finally, the classification of each point as foreground or background is based on the predicted logit scores.

The box regression network is also a PointNet \cite{qi2017pointnet} based network that relies on the foreground points as inputs and generates the 3D box parameters as outputs. Each point is processed by four layers of MLP and the output channel sizes are 128, 128, 256, and 512 respectively. Then, the last layer's output of per-point MLP (512-dim) is pooled with a max pooling layer. Finally, the max pooled features are processed by three layers (512, 256, and 39-dim) of MLP to predict the box parameters and processed by three layers (256, 256, and 1-dim) of MLP to predict the Intersection over Union (IoU) confidence scores. We parameterize the boxes similar to \cite{qi2018frustum} as the box center regression, the box heading regression and classification (\textit{i.e.} 12 heading bins) and the box size regression and classification (\textit{i.e.} 3 template sizes). To fine-tune the boxes iteratively, we utilize the same box regression network again on the foreground points that are transformed to the box coordinate as described in the main paper. In addition, we found that the cascaded box regression network performs better without sharing weights than sharing weights.

We train the autolabeler in two stages separately and adopt the losses similar to \cite{qi2018frustum}. First of all, the foreground segmentation network predicts two scores for each point that represent foreground and background, and it is supervised with a cross-entropy loss $L_{seg}$. Then, the box regression network predicts the box center regression, the box heading regression and classification, the box size regression and classification, and the IoU confidence score. We adopt 12 heading bins and 3 template sizes (3.9, 1.6, 1.56), (4.7, 2.1, 1.7), (10.0, 2.6, 3.2) that denote length, width, and height respectively. The box regression loss is defined as $L_{box_i} = w_{1,i} \times L_{h-cls_i} + w_{2,i} \times L_{h-reg_i} + w_{3,i} \times L_{s-cls_i} + w_{4,i} \times L_{s-reg_i}$ where $i \in {1, 2}$ represents the cascaded box regression network. The total loss for box regression network is defined as $L = L_{iou} + w(L_{box_1} + L_{box_2})$ where $L_{iou}$ denotes cross-entropy loss for IoU confidence score prediction. We set $w_{j,1}=1 \mbox{ for } j \in \{1,2,3,4\}$, $w_{1,2}=0.1$, $w_{2,2}=1$, $w_{3,2}=0.1$, $w_{4,2}=1$, and $w=2$ in practice.

\setlength{\tabcolsep}{0.04\linewidth}{
\begin{table*}[t]
    \centering
    \begin{tabular}{c|c|c|cc|cc}
        \specialrule{1pt}{0em}{0em}
        Task & Setting & Method & \small $\mbox{AP}_{\mbox{\scriptsize BEV}}$ & \small Closed Gap & \small $\mbox{AP}_{\mbox{\scriptsize 3D}}$ & \small Closed Gap \ts\bs\\\hline\hline
        \multirow{8}{*}{Waymo $\rightarrow$ KITTI} & - & Source Only & 88.56 & - & 86.93 & - \ts\bs\\\cline{2-7}
        & UDA & ST3D \cite{yang2021st3d} & 92.39 & +49.29\% & 92.15 & +65.99\% \ts\bs\\\cline{2-7}
        & \multirow{3}{*}{WDA} & SN \cite{wang2020train} & 86.12 & -31.40\% & 84.26 & -33.75\% \ts\\
        & & ST3D (w/ SN) \cite{yang2021st3d} & 91.37 & +36.16\% & 90.81 & +49.05\% \tts\\
        & & \cellcolor{LightCyan}\ours{} (Ours) & \cellcolor{LightCyan}\textbf{93.01} & \cellcolor{LightCyan}\textbf{+57.27\%} & \cellcolor{LightCyan}\textbf{92.96} & \cellcolor{LightCyan}\textbf{+76.23\%} \bs\\\cline{2-7}
        & - & Oracle &  96.33 & - & 94.84 & - \ts\bs\\
        \hline\hline
        \multirow{6}{*}{Waymo $\rightarrow$ nuScenes} & - & Source Only & 40.49 & - & 36.91 & - \ts\bs\\\cline{2-7}
        & UDA & ST3D \cite{yang2021st3d} & 40.91 & +2.00\% & 38.66 & +8.25\% \ts\bs\\\cline{2-7}
        & \multirow{3}{*}{WDA} & SN \cite{wang2020train} & 40.22 & -1.28\% & 36.55 & -1.70\% \ts\\
        & & ST3D (w/ SN) \cite{yang2021st3d} & 41.40 & +4.32\% & 39.02 & +9.95\% \tts\\
        & & \cellcolor{LightCyan}\ours{} (Ours) & \cellcolor{LightCyan}\textbf{42.75} & \cellcolor{LightCyan}\textbf{+10.74\%} & \cellcolor{LightCyan}\textbf{40.43} & \cellcolor{LightCyan}\textbf{+16.60\%} \bs\\\cline{2-7}
        & - & Oracle & 61.54 & - & 58.12 & - \ts\bs\\
        \hline\hline
        \multirow{6}{*}{nuScenes $\rightarrow$ KITTI} & - & Source Only & 80.73 & - & 78.51 & - \ts\bs\\\cline{2-7}
        & UDA & ST3D \cite{yang2021st3d} & 83.77 & +19.49\% & 83.65 & +31.48\% \ts\bs\\\cline{2-7}
        & \multirow{3}{*}{WDA} & SN \cite{wang2020train} & 66.72 & -89.81\% & 65.81 & -77.77\% \ts\\
        & & ST3D (w/ SN) \cite{yang2021st3d} & 90.47 & +62.44\% & 90.24 & +71.83\% \tts\\
        & & \cellcolor{LightCyan}\ours{} (Ours) & \cellcolor{LightCyan}\textbf{93.61} & \cellcolor{LightCyan}\textbf{+82.56\%} & \cellcolor{LightCyan}\textbf{93.56} & \cellcolor{LightCyan}\textbf{+92.16\%} \bs\\\cline{2-7}
        & - & Oracle & 96.33 & - & 94.84 & - \ts\bs\\
        \specialrule{1pt}{0em}{0em}
    \end{tabular} \vspace{4pt}
    \caption{\textbf{Experiment results on different DA tasks.} Our \ours{} outperforms all existing methods on $\mbox{AP}_{\mbox{\scriptsize BEV}}$ and $\mbox{AP}_{\mbox{\scriptsize 3D}}$ of the car category at \textbf{IoU = 0.5}. The reported AP are the results on the moderate case when KITTI is regarded as the target domain and are the overall results for other DA tasks. We also report the closed gap to assess how much the performance gap between Source Only and Oracle is closed.}
    \label{tab:Experiment Results at IoU = 0.5}
\end{table*}
}
\setlength{\tabcolsep}{0.035\linewidth}{
\begin{table}[t]
    \centering
    \begin{tabular}{c|c|cc}
        \specialrule{1pt}{0em}{0em}
        \multicolumn{4}{c}{\small SECOND-IoU} \ts\bs\\
        \specialrule{1pt}{0em}{0em}
        \small Setting & \small Method & \small $\mbox{AP}_{\mbox{\scriptsize BEV}}$ & \small $\mbox{AP}_{\mbox{\scriptsize 3D}}$ \ts\bs\\\hline\hline
        - & Source Only & 51.84 & 17.92 \ts\bs\\\hline
        \multirow{2}{*}{UDA} & ST3D \cite{yang2021st3d} & 75.94 & 54.13 \ts\\
        & $\mbox{ST3D++ \cite{yang2022st3d++}}^\dagger$ & 80.52 & 62.37 \bs\\\hline
        \multirow{4}{*}{WDA} & SN \cite{wang2020train} & 40.03 & 21.23 \ts\\
        & ST3D (w/ SN) \cite{yang2021st3d} & 79.02 & 62.55 \tts\\
        & $\mbox{ST3D++ (w/ SN) \cite{yang2022st3d++}}^\dagger$ & 78.87 & \textbf{65.56} \tts\\
        & \cellcolor{LightCyan}\ours{} (Ours) & \cellcolor{LightCyan}\textbf{80.66} & \cellcolor{LightCyan}64.65 \bs\\\hline
        - & Oracle & 83.29 & 73.45 \ts\bs\\
        \specialrule{1pt}{0em}{0em}
    \end{tabular}
    \caption{Experiment results on the nuScenes $\rightarrow$ KITTI task with SECOND-IoU as 3D detector. The reported $\mbox{AP}_{\mbox{\scriptsize BEV}}$ and $\mbox{AP}_{\mbox{\scriptsize 3D}}$ are the results on the moderate case of the car category at IoU = 0.7. $\dagger$~refers to the results reported by \cite{yang2022st3d++}.}
    \label{tab:Detector-agnostic Analysis}
\end{table}
}


\begin{figure}[t]
    \centering
    \includegraphics[width=\linewidth]{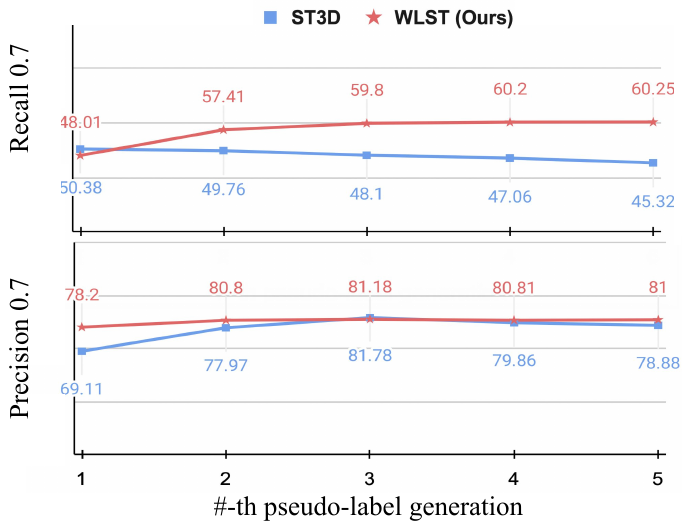}
    \caption{\textbf{Qualitative Analysis on Pseudo Labels over Time.} Comparison between our \ours{} and the state-of-the-art UDA method ST3D on the Waymo $\rightarrow$ KITTI task. We utilize Recall with IoU $>$ 0.7 and Precision with IoU $>$ 0.7 as our metrics to assess the quality of pseudo labels.}
    \label{fig:qualitative}
\end{figure}

\subsection{More Experiment Results} \label{subsec:More Experiment Results}

In this section, we present the experiment results obtained at IoU = 0.5 in Sec.~\ref{subsubsec:Experiment Results at IoU = 0.5}. We demonstrate that our \ours{} is a detector-agnostic framework in Sec.~\ref{subsubsec:Detector-agnostic Analysis} and an autolabeler-agnostic framework in Sec.~\ref{subsubsec:Autolabeler-agnostic Analysis}. Furthermore, we conduct a qualitative analysis on pseudo labels over time in Sec.~\ref{subsubsec:Qualitative Analysis on Pseudo Labels over Time} and provide a comparison with weakly-supervised methods in Sec.~\ref{subsubsec:Weakly-supervised Methods Comparison}.

\setlength{\tabcolsep}{0.035\linewidth}{
\begin{table}[t]
    \centering
    \begin{tabular}{c|c|cc}
        \specialrule{1pt}{0em}{0em}
        \multicolumn{4}{c}{\small FGR} \ts\bs\\
        \specialrule{1pt}{0em}{0em}
        \small Setting & \small Method & \small $\mbox{AP}_{\mbox{\scriptsize BEV}}$ & \small $\mbox{AP}_{\mbox{\scriptsize 3D}}$ \ts\bs\\\hline\hline
        - & Source Only & 60.32 & 21.66 \ts\bs\\\hline
        UDA & ST3D \cite{yang2021st3d} & 83.37 & 64.75 \ts\bs\\\hline
        \multirow{3}{*}{WDA} & SN \cite{wang2020train} & 78.24 & 62.54 \ts\\
        & ST3D (w/ SN) \cite{yang2021st3d} & \textbf{86.53} & \textbf{76.85} \tts\\
        & \cellcolor{LightCyan}\ours{} (Ours) & \cellcolor{LightCyan}80.47 & \cellcolor{LightCyan}69.25 \bs\\\hline
        - & Oracle & 90.85 & 83.00 \ts\bs\\
        \specialrule{1pt}{0em}{0em}
    \end{tabular} \vspace{4pt}
    \caption{Experiment results on the Waymo $\rightarrow$ KITTI task with FGR \cite{wei2021fgr} as autolabeler. The reported $\mbox{AP}_{\mbox{\scriptsize BEV}}$ and $\mbox{AP}_{\mbox{\scriptsize 3D}}$ are the results on the moderate case of the car category at IoU = 0.7.}
    \label{tab:Autolabeler-agnostic Analysis}
\end{table}
}

\subsubsection{Experiment Results at IoU = 0.5} \label{subsubsec:Experiment Results at IoU = 0.5}

As shown in Tab.~\ref{tab:Experiment Results at IoU = 0.5}, we provide the experiment results at IoU = 0.5. We observed that the baseline WDA method SN \cite{wang2020train} performs poorly with negative closed gaps by up to $\sim$-90$\%$ in $\mbox{AP}_{\mbox{\scriptsize BEV}}$ and up to $\sim$-78$\%$ in $\mbox{AP}_{\mbox{\scriptsize 3D}}$. However, our method attains a considerable performance gain and effectively closes the performance gap by up to $\sim$83$\%$ in $\mbox{AP}_{\mbox{\scriptsize BEV}}$ and up to $\sim$92$\%$ in $\mbox{AP}_{\mbox{\scriptsize 3D}}$, which outperforms current state-of-the-art WDA method ST3D (w/ SN) \cite{yang2021st3d} by up to $\sim$3$\%$ in $\mbox{AP}_{\mbox{\scriptsize BEV}}$ and up to $\sim$3$\%$ in $\mbox{AP}_{\mbox{\scriptsize 3D}}$.

\subsubsection{Detector-agnostic Analysis} \label{subsubsec:Detector-agnostic Analysis}

We demonstrate that our \ours{} is a detector-agnostic self-training framework by equipping SECOND-IoU, which is devised by \cite{yang2021st3d}, as our 3D detector. As shown in Tab.~\ref{tab:Detector-agnostic Analysis}, although our method performs less strongly than the state-of-the-art WDA method ST3D++ (w/ SN) by around $\sim$-1$\%$ on $\mbox{AP}_{\mbox{\scriptsize 3D}}$, we surpass it by around $\sim$2$\%$ on $\mbox{AP}_{\mbox{\scriptsize BEV}}$, which achieves state-of-the-art performance.

\subsubsection{Autolabeler-agnostic Analysis} \label{subsubsec:Autolabeler-agnostic Analysis}

As shown in Tab.~\ref{tab:Autolabeler-agnostic Analysis}, we demonstrate that our \ours{} is an autolabeler-agnostic self-training framework by equipping FGR \cite{wei2021fgr} as autolabeler. Since the code of the existing trainable MAP-Gen \cite{liu2022map} is not accessible, we adopt non-trainable FGR as autolabeler, which wouldn't predict IoU confidence score for each pseudo label. Hence, we only use the pseudo labels generated by FGR to check the geometric consistency and cross-modality consistency of the pseudo labels generated by 3D detector, and do not utilize them as final pseudo labels $[\hat{L}^i]_k$. Although it is a suboptimal way to select the pseudo labels, it still outperforms the current state-of-the-art UDA method ST3D \cite{yang2021st3d} by around $\sim$5$\%$ in $\mbox{AP}_{\mbox{\scriptsize 3D}}$, which validates that our \ourfusionstrategy{} can effectively improve the quality of pseudo labels in accordance with geometric consistency and cross-modality consistency.

\subsubsection{Qualitative Analysis on Pseudo Labels over Time} \label{subsubsec:Qualitative Analysis on Pseudo Labels over Time}

To validate that our \ours{} framework can generate more robust and consistent pseudo labels than the UDA approaches, we further conduct the qualitative analysis on pseudo labels in the pseudo-label generation process over time as shown in Fig.~\ref{fig:qualitative}. In the first pseudo-label generation iteration, despite having lower recall than ST3D, ours has relatively high precision that ensures the model is less likely to be affected by noisy pseudo labels. Then, our recall and precision continue to improve stably over time while the recall of ST3D drops incrementally, suggesting that precision is more crucial than recall on the initial pseudo labels.

\subsubsection{Weakly-supervised Methods Comparison} \label{subsubsec:Weakly-supervised Methods Comparison}

As shown in Tab.~\ref{tab:Weakly-supervised Methods Comparison}, we compare our method \ours{} to existing weakly-supervised 3D object detection methods to validate the significance of our WDA method. While we didn't conduct a direct comparison with weakly-supervised methods, prior work \cite{wang2020train, yang2021st3d, yang2022st3d++} has shown the limitations (poor performance) of applying learning-based weakly-supervised 3D object detection methods directly to new environments without utilizing information or statistics from target domain 3D bounding boxes. Despite that, we conducted additional experiments to investigate the performance of learning-based methods when they were trained on the target domain and inevitably utilized a portion of 3D labels. For instance, MAP-Gen utilizes 500 frames out of a total of 3712 frames and WS3D (2020, 2021) utilizes 534 precisely annotated 3D labels to train their autolabeler. In contrast, the non-learning-based method FGR and our WDA method \ours{} do NOT require any 3D labels. Even though this comparison may be considered unfair for FGR and our method \ours{}, it is noteworthy that our method still outperforms WS3D (2020, 2021) by around $\sim$2$\%$ in $\mbox{AP}_{\mbox{\scriptsize 3D}}$ for the moderate case and achieves comparable performance to MAP-Gen, which utilizes around 13.5$\%$ of 3D labels. Moreover, our method significantly outperforms the non-learning-based method FGR, particularly in the hard case by around $\sim$8$\%$ in $\mbox{AP}_{\mbox{\scriptsize 3D}}$. Overall, these results provide strong validation for the significance of our WDA method.

\setlength{\tabcolsep}{0.01\linewidth}{
\begin{table}[t]
    \centering
    \begin{tabular}{l|c|c|ccc}
        \specialrule{1pt}{0em}{0em}
        \multirow{2}{*}{\small Method} & \multicolumn{2}{c}{\small Target Domain}\vline & \multicolumn{3}{c}{\small $\mbox{AP}_{\mbox{\scriptsize 3D}}$} \ts\bs\\\cline{2-6}
        & \footnotesize Weak label & \footnotesize 3D label & \footnotesize Easy & \footnotesize Moderate & \footnotesize Hard \ts\bs\\\hline\hline
        \small WS3D \footnotesize (2020) \small \cite{meng2020weakly} & \footnotesize BEV Centroid & \checkmark & \small 84.04 & \small 75.10 & \small 73.29 \ts\\
        \small WS3D \footnotesize (2021) \small \cite{meng2021towards} & \footnotesize BEV Centroid & \checkmark & \small 85.04 & \small 75.94 & \small 74.38 \tts\\
        \small MAP-Gen \small \cite{liu2022map} & \footnotesize 2D box & \checkmark & \small 87.87 & \textbf{\small 77.98} & \textbf{\small 76.18} \tts\\
        \small FGR \small \cite{wei2021fgr} & \footnotesize 2D box & & \small 86.68 & \small 73.55 & \small 67.91 \bs\\\hline
        \small \cellcolor{LightCyan}\ours{} (Ours) & \cellcolor{LightCyan}\footnotesize 2D box & \cellcolor{LightCyan} & \small \cellcolor{LightCyan}\textbf{88.01} & \small \cellcolor{LightCyan}77.31 & \small \cellcolor{LightCyan}76.03 \ts\bs\\
        \specialrule{1pt}{0em}{0em}
    \end{tabular} \vspace{6pt}
    \caption{We compare our WDA method \ours{} to existing weakly-supervised 3D object detection methods and report $\mbox{AP}_{\mbox{\scriptsize 3D}}$ of the car category at IoU = 0.7 on the nuScenes $\rightarrow$ KITTI task.}
    \label{tab:Weakly-supervised Methods Comparison}
\end{table}
}

\end{document}